\documentclass[11pt,a4paper]{article}
\usepackage[hyperref]{emnlp2020}
\usepackage{times}
\usepackage{latexsym}
\usepackage{amsmath}
\usepackage{bbm}
\usepackage{tabularx}
\usepackage{booktabs}

\usepackage{graphicx}

\usepackage{amsthm}
\theoremstyle{definition}
\newtheorem{definition}{Definition}[section]

\usepackage{microtype}

\aclfinalcopy 

\title{BLEU Neighbors: \\ A Reference-less Approach to Automatic Evaluation}

\author{Kawin Ethayarajh \\
  Stanford University \\
  \texttt{kawin@stanford.edu} \\\ \And
  Dorsa Sadigh \\
  ILIAD Lab, Stanford University \\
  \texttt{dorsa@cs.stanford.edu}
  }

\date{}

\begin{document}
\maketitle
\begin{abstract}
Evaluation is a bottleneck in the development of natural language generation (NLG) models. Automatic metrics such as BLEU rely on references, but for tasks such as open-ended generation, there are no references to draw upon. Although language diversity can be estimated using statistical measures such as perplexity, measuring language \emph{quality} requires human evaluation. However, because human evaluation at scale is slow and expensive, it is used sparingly; it cannot be used to rapidly iterate on NLG models, in the way BLEU is used for machine translation. To this end, we propose \emph{BLEU Neighbors}, a nearest neighbors model for estimating language quality by using the BLEU score as a kernel function. On existing datasets for chitchat dialogue and open-ended sentence generation, we find that -- on average -- the quality estimation from a BLEU Neighbors model has a lower mean squared error and higher Spearman correlation with the ground truth than individual human annotators. Despite its simplicity, BLEU Neighbors even outperforms state-of-the-art models on automatically grading essays, including models that have access to a gold-standard reference essay. 
\end{abstract}

\section{Introduction}
\label{intro}

\begin{figure}
\centering
\includegraphics[width=1.05\columnwidth]{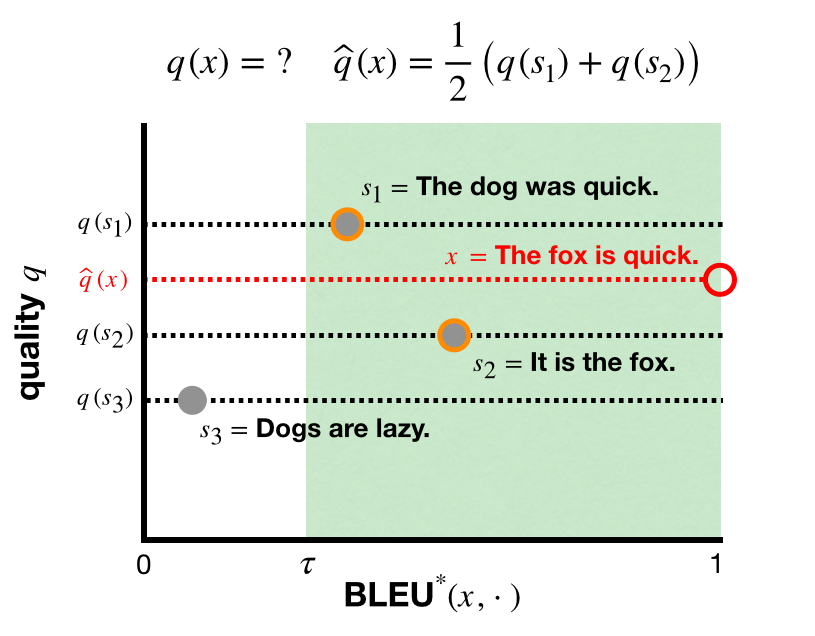}
\caption{We want to score a sentence $x$ given training examples $S = \{s_1, s_2, s_3\}$ with known quality scores $\{ q(s_1), q(s_2), q(s_3)\}$. BLEU Neighbors works as follows: calculate $\texttt{BLEU}^*(x, \cdot)$, a variant of the BLEU-4 score, for each $s$; ignore those below $\tau = 0.08$; take the average score of those that remain to predict $\widehat{q}(x)$.}
\label{teaser}
\end{figure}

Despite recent advances on many natural language generation (NLG) tasks -- including open-ended generation, chitchat dialogue, and abstractive summarization -- evaluation remains a challenge. Automatic metrics such as BLEU rely on references, but for many NLG tasks, there is no single correct answer. In dialogue, the space of acceptable responses to a given prompt is often very large, yet most datasets only provide a few gold-standard references \citep{serban2015survey}. In open-ended generation, where text is generated freely by a language model, there are no references at all; statistical measures such as perplexity capture language diversity but not language quality \citep{hashimoto2019unifying}. These limitations necessitate human evaluation. However, because human evaluation at scale is slow and expensive, it is used sparingly; it cannot be used to rapidly iterate on NLG models, in the way BLEU is used for machine translation. 

Prior work on automating reference-less evaluation has largely been limited in scope. Heuristic-based evaluation was found to be effective for grammatical error correction, but the methods used were problem-specific and cannot be extended to other tasks \citep{napoles2016there,choshen2018reference,asano2017reference}. Using the log-odds from a language model, \citet{kann2018sentence} made automatic judgments of sentence-level fluency that correlated moderately well with human judgment, but this captured only one facet of language quality. Approaches that were broader in scope found less success: although ADEM, an RNN trained to score dialogue responses, was initially thought to correlate well with human judgment \citep{lowe2017towards}, it was later found to generalize poorly, placing outsized influence on factors such as response length \citep{retrospective}.

Can we come up with a fast and simple method for reference-less evaluation of language quality, analogous to BLEU for machine translation? Note that our goal here is \emph{not} to supplant human evaluation, but to complement it: as long as the method's predictions correlate moderately well with the ground-truth quality scores, it can be used to speed up NLG model development. Our desiderata are then as follows: simplicity, speed, and a moderately strong correlation with the ground truth. To this end, we propose \emph{BLEU Neighbors}, a new approach to reference-less automatic evaluation.

Our approach is a nearest neighbors model that predicts language quality by using BLEU as a kernel function. We start with training examples $S$, where each sentence $s \in S$ has a ground-truth quality score $q(s)$. Note that these examples are not references -- we do not expect the NLG model being evaluated to generate any sentence in $S$. In fact, $S$ contains sentences of varying quality, including incoherent sentences with low quality scores. Given a test sentence $x$, we use the BLEU score to identify its neighbors in the training data: $\{ s\ |\ \texttt{BLEU}^*(x,s) > \tau, s \in S \}$, where $\tau$ is a similarity threshold. Then we simply take the mean of the neighbors' known quality scores to estimate $\widehat{q}(x)$, the quality of $x$. Consider the test sentence $x = \textit{`The fox is quick'}$. As seen in Figure \ref{teaser}, it overlaps with $s_1 = \textit{`The dog was quick.'}$ and $s_2 = \textit{`It is the fox.'}$ but not with $s_3 = \textit{`Dogs are lazy.'}$. Therefore, we estimate $\widehat{q}(x)$ as the mean of $q(s_1)$ and $q(s_2)$ but not $q(s_3)$. 

We test BLEU Neighbors on the datasets from HUSE \citep{hashimoto2019unifying}, where each sentence's ground-truth quality score is the average over 20 human judgments. On the dialogue and open-ended generation datasets, we find that -- on average -- the BLEU Neighbors model has a lower mean squared error (MSE) and higher Spearman correlation with the ground truth than individual annotators. The premise of our method is that past approaches to reference-less evaluation fell short because they were too ambitious -- if a given test sentence is not sufficiently similar to any training example, no prediction should be made at all. Although we sacrifice some coverage in order to make more accurate estimates, this sacrifice is modest: BLEU Neighbors makes predictions for 41\% to 99\% of sentences from the HUSE datasets. Our method is also data-efficient -- none of HUSE datasets have over 400 training examples.

Our method is weakest on evaluating summaries; this is unsurprising, given that summary quality is conditioned on the source text, which the method ignores. In contrast, BLEU Neighbors is surprisingly effective at automatically grading essays, achieving a new state-of-the-art and even beating out models that have access to a gold-standard reference essay. These findings suggest that despite its simplicity, our approach has broad applicability. Although BLEU Neighbors does not measure language \emph{diversity}, it is sufficient for it to measure quality alone. The former is easier to estimate (e.g., perplexity) and can be combined with the BLEU Neighbors score in a hybrid metric \citep{hashimoto2019unifying}. We conclude by providing some practical advice, such as how to prevent NLG models from explicitly optimizing for a high BLEU Neighbors score without generating high-quality output.

\section{Related Work}

\subsection{Reference-based Evaluation}

BLEU \citep{papineni2002bleu}, ROUGE \citep{lin-2004-rouge}, and METEOR \citep{banerjee2005meteor} are the \emph{de facto} canonical metrics of reference-based automatic evaluation. Given a candidate sentence $x$ and a reference sentence $s$, each metric assigns a score $q(x,s) \in [0,1]$ based on how well $x$ overlaps with $s$. Where the metrics differ is in how they define this overlap. Letting $\ell_n(\cdot)$ denote the list of $n$-grams, the $n$-gram precision $P_n$ and recall $R_n$ can be defined as follows:
\begin{equation}
    \begin{split}
        P_n(x,s) & = \frac{1}{|\ell_n(x)|} \sum_{g \in \ell_n(x)} \mathbbm{1}[g \in \ell_n(s)]  \\
        R_n(x,s) & = \frac{1}{|\ell_n(s)|} \sum_{g \in \ell_n(s)} \mathbbm{1}[g \in \ell_n(x)]  \\
    \end{split}
    \label{eq:precision_and_recall}
\end{equation}

\paragraph{BLEU} The BLEU score for $(x,s)$ is the geometric mean of the $n$-gram precision $P_n$ up to a chosen $n$ (typically, $n = 4$). BLEU also implements clipping, such that each $n$-gram $g \in \ell_n(x)$ can be matched at most once. It also includes a brevity penalty to penalize shorter candidates.

\paragraph{METEOR} The METEOR score takes the harmonic mean of $P_1$ and $R_1$, with greater weight placed on $R_1$. It is laxer than BLEU, allowing words in $x$ and $s$ to match, for example, if they are synonyms or share the same stem \citep{banerjee2005meteor}. Instead of looking at higher order $n$-grams, METEOR tries to align the tokens in $x$ and $s$ and penalizes alignments that are not contiguous.

\paragraph{ROUGE-L} ROUGE-L, the variant of ROUGE we discuss in this work, measures the overlap between $x$ and $s$ as the size of their longest common subsequence $\textit{LCS}(x,s)$. Specifically, it calculates $\textit{LCS}(x,s) / P_1(x,s)$ and $\textit{LCS}(x,s) / R_1(x,s)$ and takes their harmonic mean.

Although there have been advances in reference-based automatic evaluation -- such as BEER \citep{stanojevic2014beer} and RUSE \citep{vedantam2015cider}, among others \citep{shimanaka2018ruse,ma2017blend,lo2018accurate,zhao2019moverscore} -- BLEU and METEOR are still widely used for machine translation; ROUGE, for summarization \citep{liu2016not}. This is partially because some of the newer methods are learned metrics that do not generalize well to new domains \citep{chaganty2018price}. Moreover, most do not enjoy the incumbent status that BLEU, ROUGE, and METEOR have. To our knowledge, the current state-of-the-art in reference-based evaluation metrics is BERTScore \citep{zhang2019bertscore}, which uses BERT embeddings \citep{devlin2019bert} to compute similarity at the token-level before aggregating the similarities using importance-weighting. As it is state-of-the-art for reference-based evaluation, it is the only non-canonical metric we consider as a kernel function.

\subsection{Reference-less Evaluation}

Compared to reference-based evaluation, little work has been done on automating reference-less evaluation. The most successful approaches have been task-specific: heuristic-based evaluation was found to be effective for grammatical error correction \citep{napoles2016there,choshen2018reference,asano2017reference}. However, those heuristics cannot be extended to other tasks. \citet{kann2018sentence} proposed two metrics for judging the fluency of a sentence: sentence-level log-odds ratio (SLOR) and a Wordpiece-based variant named WPSLOR. Although the latter correlates moderately well (Pearson's $r > 0.40$) with human judgment, it should be noted that sentence-level fluency is only one facet of language quality --  a sentence may be probable according to a language model while making little sense to a human.

Approaches that were broader in scope were less successful. ADEM, an RNN trained to score dialogue responses, was initially thought to correlate well with human judgment \citep{lowe2017towards}. However, the authors later found that it generalized poorly \citep{retrospective}, placing outsized influence on factors such as response length. It was also found to be vulnerable to adversarial examples \citep{sai2019re}. In any case, ADEM was not a purely reference-less method -- it still required a gold-standard reference as input. Rather, its key insight was that the space of acceptable responses is much larger than the handful of gold-standard references provided in dialogue datasets, and that this should be considered when estimating quality.

\section{BLEU Neighbors}
\label{sec:model}

Given a candidate sentence $x$, training examples $S$, and ground-truth quality scores $\{ q(s)\ |\ s \in S\}$, we want to estimate $\widehat{q}(x)$, the language quality of $x$. How can we do so in a fast and simple manner such that our predictions correlate well with the ground truth? We propose a nearest neighbors model that uses a variant of the BLEU score called $\text{BLEU}^*$ as the kernel function. Once the neighbors of $x$ have been identified, we take the mean of their known quality scores as $\widehat{q}(x)$.

\begin{definition}
The \emph{non-unigram BLEU-4 score} is a variant of the BLEU-4 score that ignores unigram precision.  Where $\beta = \exp \left( \min \left( 0, 1 - \frac{|\ell_1(s)|}{|\ell_1(x)|} \right) \right)$ is the brevity penalty and $P_i$ is defined in (\ref{eq:precision_and_recall}),
\begin{equation}
    \begin{split}
        \text{BLEU}^*(x,s) &= \beta \cdot \prod_{i = 2}^{4} {P_i(x,s)}^{1/3} \\
    \end{split}
\end{equation}
BLEU Neighbors uses this variant of BLEU as the kernel function. We ignore the unigram precision $P_1$ because we are not comparing candidates and their direct references, but rather candidates and training examples. It is not uncommon for two random sentences to have stopwords in common, in which case a non-zero $P_1$ is unexceptional. We validated this empirically as well, finding that ignoring $P_1$ improves correlation with the ground-truth.
\end{definition}

\begin{definition}
    Given a candidate sentence $x$, training examples $S$, and a similarity threshold $\tau \in [0,1]$ , the \emph{BLEU neighbors} of $x$ are $$\mathcal{N} = \{ s \in S\ |\ \text{BLEU}^*(x,s) \geq \tau \}$$
    To ensure that the quality estimate is stable, we require that $\mathcal{N}$ have a minimum size of $a \in \mathbbm{Z}^+$. Conversely,  a candidate sentence that overlaps with many training examples in $S$ likely does so because it contains many common $n$-grams. This complicates evaluation: since $\text{BLEU}^*$ does not weigh $n$-grams by their frequency, an abundance of common $n$-grams -- such as \emph{``on the''} or \emph{``it is''}, for example -- can exaggerate the similarity between the candidate and a training example. In this scenario, it is best that no prediction be made at all. Since $\mathcal{N} \subseteq \mathcal{S}$, let $b \in [0,1]$ denote the largest fraction of $S$ that $\mathcal{N}$ can contain. We express $b$ as a fraction of the training set size $|S|$ because if $S$ is very large, it would not be uncommon for even sentences with rare $n$-grams to have matches in $S$. 
    
    When $\mathcal{N}$ meets the aforementioned size constraints, the BLEU Neighbors estimate of $x$'s quality is the average of its neighbors' quality scores:
    \begin{equation}
    \begin{split}
        \widehat{q}(x) &= 
        \left\{\;
        \begin{split}
            \frac{1}{|\mathcal{N}|} \sum_{s \in \mathcal{N}} q(s) & \qquad a \leq |\mathcal{N}| \leq b |S| \\
            \text{undefined} & \qquad \text{otherwise}\\
        \end{split}
        \right. \\
    \end{split}
    \end{equation}
    In other words, $\mathcal{N} \subseteq S$ comprises all the training examples that are sufficiently similar to the candidate with respect to $\text{BLEU}^*$. If there are fewer than $a$ examples or more than $b|S|$ examples in $\mathcal{N}$, then no prediction is made; otherwise, the estimate $\widehat{q}(x)$ is the average quality of the examples in $\mathcal{N}$.
    \label{def:evidence_thresholds}
\end{definition}

Although $\tau, a, b$ are parameters to be set, we find that $\tau = 0.08, a = 5, b = 0.66$ are near-optimal for all tasks (see section \ref{ssec:varying}). This universality allows BLEU Neighbors to be used out-of-the-box, without hyperparameter tuning. Note that $S$ should only be used to train the evaluator (i.e., BLEU Neighbors). The generator (i.e., the NLG model being evaluated) should not have access to $S$; otherwise, it could optimize for a high BLEU Neighbors score by including $n$-grams that only belong to examples with a high ground-truth quality, thus artificially inflating the quality estimates. 

\begin{definition}
    Given a set of candidates $\mathcal{X}$ to be evaluated, the \emph{coverage} of $\mathcal{X}$ is the proportion of candidates for which $\widehat{q}(x)$ is defined.
\end{definition}

This is a key distinction between BLEU Neighbors and prior approaches to reference-less evaluation: our approach does not necessarily make a prediction for all candidates. This is by design -- as mentioned earlier, we surmise that past approaches fell short because they were too ambitious, trying to score sentences that simply could not be scored. There is a trade-off between coverage and prediction error, with greater coverage generally coming at the cost of greater prediction error.

\begin{table*}[t]
    \centering
    \small
    \begin{tabularx}{\textwidth}{Xccc|ccc|ccc}
\toprule
 & \multicolumn{3}{c|}{\textbf{Dialogue}} & \multicolumn{3}{c|}{\textbf{Open-ended Generation}} & 
\multicolumn{3}{c}{\textbf{Summarization}} \vspace{0.1cm} \\

{} &  MSE & $\rho$ & Coverage & MSE & $\rho$ & Coverage &  MSE & $\rho$ & Coverage \\
\midrule

Human (best)            &                              0.0208 &                             0.878\ \ \ &                                     1.00 &              0.0177 &             0.861\ \ \ &                     1.00 &                                        0.0200 &                                      0.921\ \ \ &                                               1.00 \\
Human (average)              &                              0.0807 &                                     0.456\ \ \ &                                     1.00 &              0.0719 &                     0.472\ \ \ &                     1.00 &                                        0.0802 &                                              0.405\ \ \ &                                               1.00 \\ \midrule

BLEU Neighbors            &                              \bf 0.0164 &                            \bf 0.470* &                                     0.76 &              0.0204 &             \bf 0.575* &                     0.41 &                                        0.0213 &                                      \bf 0.325* &                                               0.99 \\
ROUGE Neighbors           &                              0.0197 &                             0.342* &                                     0.86 &             \bf 0.0174 &             0.077\ \ \ &                     0.47 &                                        0.0226 &                                      0.245* &                                               0.97 \\
METEOR Neighbors          &                              0.0165 &                             0.382* &                                     0.47 &              0.0209 &             0.395\ \ \ &                     0.22 &                                        \bf 0.0180 &                                      0.240\ \ \ &                                               0.12 \\
BERTScore Neighbors            &                              0.0229 &                             0.150* &                                     0.89 &              0.0192 &             0.566* &                     0.32 &                                        0.0223 &                                      0.225\ \ \ &                                               0.53 \\
\bottomrule

\end{tabularx}

    \caption{The mean squared error (MSE) and Spearman's $\rho$ of language quality predictions $\widehat{q}(\cdot)$ with respect to the ground truth $q(\cdot)$. The lowest MSE and highest $\rho$ across all models is in bold and * signifies $p < 0.01$. For all tasks, BLEU Neighbors achieves a higher Spearman's $\rho$ than its ROUGE, METEOR, and BERTScore counterparts. For dialogue and open-ended generation, it even has a lower MSE and higher $\rho$ than human annotators on average.}
    \label{tab:huse}
\end{table*}

\section{Experiments}
\label{sec:experiments}

\paragraph{NLG Tasks} We test BLEU Neighbors on evaluating sentences from the following NLG tasks: chitchat dialogue, open-ended sentence generation (from a language model), and abstractive summarization. \citet{hashimoto2019unifying} provided a dataset for each of these tasks, which we collectively refer to as the HUSE datasets. We ignore the story generation dataset in that work because the machine-generated examples are far from human quality and can thus be trivially assigned a low quality score.

Each dataset contains a mixture of machine- and human-generated sentences, in roughly equal proportion. Each sentence in the HUSE datasets was judged by 20 human annotators, who assigned it a label based on its typicality. These labels map to an integer score from 0 to 5. We divide the raw judgment by 5 to bound it in $[0,1]$ and then take the mean across all 20 annotators, which we treat as the ground-truth language quality $q(s)$ for each sentence $s$. Because these datasets are small, we use leave-one-out prediction. That is, given a candidate sentence from a particular HUSE dataset, we treat the remaining $n - 1$ sentences as $S$.

\paragraph{Grading Essays} We also test our model on automatically grading essays from the ASAP-SAS dataset\footnote{https://www.kaggle.com/c/asap-sas}. Although each essay is a multi-sentence paragraph, we did not adapt our model in any way. Each essay's quality score is an integer from 0 to 3, which we divide by 3 to bound in $[0,1]$. This normalization is done for the sake of consistency. Because there are distinct training and test sets, we draw the training examples from the training data and the candidates to be evaluated from the test data. The ASAP-SAS data is also broken down by topic. The current state-of-the-art model only evaluates on topic \#3 -- specifically, on essays from topic \#3 that contain 5 to 15 sentences \citep{clark2019sentence}. Therefore, to allow for a fair comparison, we also draw test sentences from this subset. 

\begin{figure*}
\centering
\includegraphics[width=\textwidth]{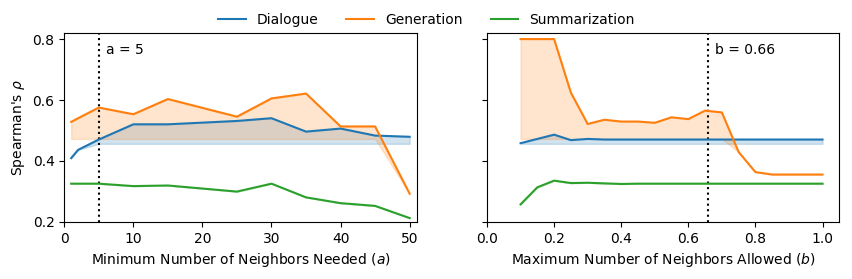}
\caption{Spearman's $\rho$ between BLEU Neighbors estimates $\widehat{q}(\cdot)$ and the ground-truth quality $q(\cdot)$ as each evidence threshold changes, while the other is held constant at $a = 5$ or $b = 0.66$. $a$ is the minimum number of neighbors needed; $b$ is the maximum allowed (as a fraction of the training set). For all tasks, increasing $a$ improves correlation, up to a point. Only the correlation for open-ended generation is sensitive to changes in $b$, which decreases as $b$ increases. The shaded area for each task indicates above-human performance (on average).}
\label{fig:spearmanr}
\end{figure*}

\paragraph{Threshold Settings} Unless otherwise stated, for all HUSE datasets, we use $\tau = 0.08, a = 5, b = 0.66$. These settings were chosen to maximize the Spearman correlation with the ground-truth quality while retaining at least 40\% coverage. The same settings were used for essay grading, except with no upper bound on $|\mathcal{N}|$ (i.e., $b = 1$), since each essay is a multi-sentence text that has some overlap with most essays in the training data. In section \ref{ssec:varying}, we show how $a, b$ can be adjusted to trade off some performance for greater coverage (and vice-versa).

\paragraph{Other Kernel Functions} In addition to using a variant of the BLEU score as the kernel function, we try other automatic metrics, including ROUGE, METEOR, and BERTScore \citep{zhang2019bertscore}. As with BLEU, a single value of $\tau$ for each metric works universally well: 0.06 (for ROUGE); 0.18 (for METEOR); 0.10 (for BERTScore).

\section{Results}
\label{sec:results}

\subsection{BLEU Neighbors vs. Humans}

In Table \ref{tab:huse}, using mean squared error (MSE) and the Spearman correlation, we compare the language quality predictions $\widehat{q}(\cdot)$ made by our various models with the ground-truth quality $q(\cdot)$. Because the ground-truth quality is the mean over 20 annotator judgments, we provide the performance of the best human annotator and the average performance across all individual annotators. Note that not all annotators scored all the examples: the average MSE and $\rho$ we report in Table \ref{tab:huse} is the average over what each annotator obtained on their respective subset of the data. We find that there is a significant gap between the best- and average-case, both in terms of MSE and Spearman's $\rho$. For example, on the summarization task, the MSE and Spearman's $\rho$ of the best human annnotator is 4x and 2x better than those of annotators on average.

\begin{figure*}
\centering
\includegraphics[width=\textwidth]{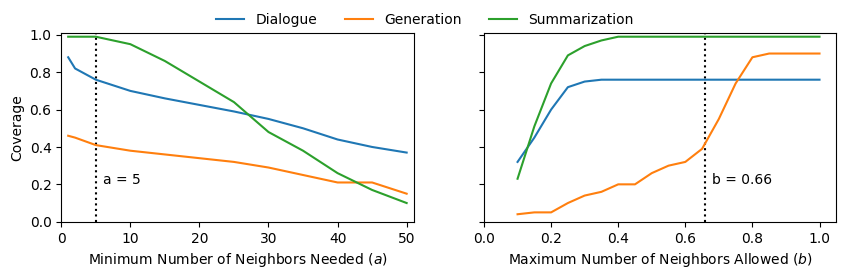}
\caption{The coverage (i.e., the fraction of sentences for which BLEU Neighbors makes predictions) as each evidence threshold changes while the other is held constant at $a = 5$ or $b = 0.66$. $a$ is the minimum number of neighbors needed; $b$ is the maximum number allowed (as a fraction of the training set). For all tasks, coverage falls as $a$ increases and $b$ decreases (i.e., as the range for the acceptable number of neighbors gets smaller).}
\label{fig:coverage}
\end{figure*}

\paragraph{Spearman Correlation} As shown in the second section of Table \ref{tab:huse}, for all tasks, we find that BLEU Neighbors has a higher Spearman correlation with the ground truth than its ROUGE, METEOR, and BERTScore counterparts. For open-ended generation and dialogue, it even outperforms the average-case human annotator. Only on evaluating summaries does the average-case annotator beat all nearest neighbors models with respect to Spearman's $\rho$; this is unsurprising, given that summary quality is strongly conditioned on the source text, which these models ignore. Despite the impressive performance of BLEU Neighbors, it should be noted that it is still well behind the \emph{best} human annotator for each task.

\paragraph{Mean Squared Error} Although BLEU Neighbors achieves a much higher Spearman correlation with the ground-truth quality than its counterparts, the model that achieves the lowest MSE varies across tasks. How can we reconcile these observations? We find that the variance of the ground-truth quality is quite small for all datasets. By just predicting the mean of $q(\cdot)$ for all candidates, we can get an MSE for each task that is only slightly higher than the best annotator's. Models that obtain the lowest MSE while also having a low Spearman's $\rho$ are thus making low-variance estimates close to the mean that do not correlate well with the ground truth. Also, annotators of the HUSE datasets assigned discrete scores \citep{hashimoto2019unifying}, while $q(\cdot)$, being an average over those scores, is continuous. This is conducive to human annotators having a higher MSE than the models.

\subsection{Varying the Evidence Thresholds}
\label{ssec:varying}

Recall that BLEU Neighbors has two evidence thresholds: $a$, the minimum number of neighbors needed to make a prediction, and $b$, the maximum number of neighbors allowed (as a fraction of the training set $S$). In Figure \ref{fig:spearmanr}, we plot the Spearman correlation between predictions $\widehat{q}(\cdot)$ and the ground truth $q(\cdot)$ as each threshold changes, while the other is held constant at the default setting ($a = 5, b = 0.66$). In Figure $\ref{fig:coverage}$, we plot the change in coverage as the thresholds change.

\paragraph{Spearman Correlation} The correlation for all tasks is sensitive to changes in $a$, with the correlation peaking at $a = 30$ or $a = 35$ before declining. This is intuitive: increasing the amount of evidence required yields more robust predictions, but sentences that meet the stringent requirement of having at least $a \geq 35$ neighbors likely have many common $n$-grams, making them harder to score. While performance on all tasks is sensitive to $a$, only performance on open-ended generation is sensitive to $b$, with the correlation decreasing as $b$ increases (i.e., as we loosen the upper bound on the number of neighbors). This suggests that sentences in the dialogue and summarization datasets do not have many neighbors to begin with, which is why tightening the upper bound has little effect. Sentences in the open-ended generation data, on the other hand, seem to have many more neighbors on average, resulting in $\rho$ being inversely related to $b$. The two sudden drops in Spearman's $\rho$ for open-ended generation -- at approximately $b = 0.2$ and $b = 0.7$ -- suggests that the distribution of $|\mathcal{N}|$, the number of neighbors, is multi-modal.

\paragraph{Coverage within a Model} The higher $a$ is and the lower $b$ is, the more candidates we reject for having too few or too many neighbors. In Figure \ref{fig:coverage}, the coverage falls linearly as $a$ increases but rises linearly before plateauing as $b$ increases. The plateau is indicative of no candidate sentence having that many neighbors to begin with.

\paragraph{Coverage across Models} Holding constant the evidence thresholds $a$ and $b$, we see in Table \ref{tab:huse} that coverage \emph{across} different models is unrelated to MSE and Spearman's $\rho$. For all models, $\tau$ is set to minimize the MSE and maximize Spearman's $\rho$. However, models with the lowest MSE or highest $\rho$ on a given task are not necessarily the most selective (i.e., those with the lowest coverage). BLEU Neighbors, which has the highest correlation on all tasks, has a coverage of 41\%, 76\%, and 99\% on open-ended generation, dialogue, and summarization respectively. In other words, the trade-off between coverage and prediction error exists \emph{within} a model -- as a function of parameters $a$ and $b$ -- but not \emph{across} different types of models.

\paragraph{Performance vs. Coverage} As seen in Figures \ref{fig:spearmanr} and \ref{fig:coverage}, there is a trade-off when choosing $a$ and $b$. Higher $a$ and lower $b$ result in better performance (i.e., greater correlation with the ground truth), but they also decrease coverage. Recall the default settings: $a = 5, b = 0.66$. Even though correlation on most tasks peaks at $a = 30$ or $a = 35$, we choose $a = 5$ as the default because we want to keep the coverage as high as possible. By choosing $a > 1$, however, we still see some benefit from requiring a minimum number of neighbors. We choose $b = 0.66$ because it is near the end of a plateau past which performance on open-ended generation data drops precipitously. In other words, the default settings of $a,b$ are near Pareto-optimal, maximizing coverage while outperforming human annotators on average. Some performance can be traded off for additional coverage by picking a different point $(a,b)$ on the Pareto frontier.

\begin{table}
    \small
    \centering
    \begin{tabularx}{\columnwidth}{X|ccc}
    \toprule
        \multicolumn{1}{c}{Source Task} & \multicolumn{3}{c}{Target Task} \\ \midrule
         &  $\rightarrow$ D & $\rightarrow$ G  & $\rightarrow$ S \\
         Dialogue (D) $\rightarrow$ & \bf 0.470 & 0.206 & 0.032 \\
         Generation (G) $\rightarrow$ & 0.310 & \bf 0.575 & -0.070 \\
         Summarization (S) $\rightarrow$ & 0.276 & 0.095 & \bf 0.325 \\ \bottomrule
    \end{tabularx}
    \caption{BLEU Neighbors performance when the training and test examples are sourced from different tasks. For example, the intersection of \emph{G $\rightarrow$} and \emph{$\rightarrow$ D} means that training examples from open-ended generation are used to score dialogue data. In this setup, a moderate Spearman's $\rho$ can still be achieved on the dialogue data.}
    \label{tab:cross_task}
\end{table}

\subsection{Low-Hanging or High-Hanging Fruit?}

Does BLEU Neighbors only make predictions for sentences that humans consider easy to score (i.e., low-hanging fruit)? Let $\mathcal{A}_i$ denote the set of all sentences for task $i$ and $\mathcal{B}_i \subseteq \mathcal{A}_i$ denote the subset of those sentences for which BLEU Neighbors makes predictions. We can answer this question by comparing the average MSE of human annotators on $\mathcal{A}_i$ with their average MSE on $\mathcal{B}_i$, which we will denote as $\overline{\text{MSE}}(\mathcal{A}_i)$ and $\overline{\text{MSE}}(\mathcal{B}_i)$ respectively. We cannot use the Spearman correlation for comparison because not every annotator scored every sentence; recall that the statistics reported in Table \ref{tab:huse} are computed over each annotator's performance on their subset of the data.

If our model were only scoring the easy-to-score sentences, then we would expect $\overline{\text{MSE}}(\mathcal{A}_i)$ to be significantly larger than $\overline{\text{MSE}}(\mathcal{B}_i)$. However, for both summarization and open-ended generation, we find that there is no statistically significant difference between these means at any level. Only on the dialogue dataset could this theory partially explain the success of our model: $\overline{\text{MSE}}(\mathcal{B}_{\text{dialogue}})$ is 15.6\% lower than $\overline{\text{MSE}}(\mathcal{A}_{\text{dialogue}})$ and this difference is significant at $p < 0.01$. However, the average-case annotator MSE on the subset of the dialogue data scored by ROUGE Neighbors is only $2 \times 10^{-4}$ higher than $\overline{\text{MSE}}(\mathcal{B}_{\text{dialogue}})$, yet ROUGE Neighbors performs far worse than its BLEU counterpart (see Table \ref{tab:huse}). This implies that the success of BLEU Neighbors is much more than simply picking the right sentences to score.

\subsection{Cross-Task Performance}

In Table \ref{tab:cross_task}, we report the Spearman's $\rho$ for BLEU Neighbors when the training and test examples are drawn from different tasks. Of all the tasks, performance on dialogue is the most robust: regardless of which task is used to source the training data, it is possible to achieve a moderately strong correlation ($\rho > 0.27$), albeit with lower coverage. Performance on summarization drops to near zero in this setup -- this is unsurprising, given that summary quality is strongly conditioned on the source text, which is ignored. For open-ended generation, it is still possible to achieve a weak correlation ($\rho > 0.09$) with this setup. Curiously, the coverage for open-ended generation actually improves when the training data is sourced from a different task, so it may be possible to adjust parameters $a,b$ to trade off some coverage for a higher correlation.

\begin{figure}[t]
    \centering
    \includegraphics[width=\columnwidth]{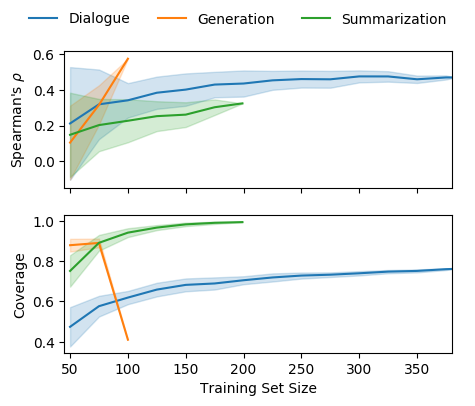}
    \caption{BLEU Neighbors performance when only a random subset of the training data is used. With more data, both coverage and the Spearman correlation with the ground truth improve, albeit with diminishing returns. The shaded area denotes one standard deviation (i.e., variation in performance across random samples).}
    \label{fig:training_size}
\end{figure}

\subsection{How much Training Data is Needed?}

In Figure \ref{fig:training_size}, we plot the performance of BLEU Neighbors on the HUSE datasets for different amounts of training data. This is simulated by drawing a random subset of the data with $n$ examples, doing hold-one-out prediction with $n - 1$ examples, and then taking the mean performance over 20 such runs. We find that BLEU Neighbors is surprisingly robust on all tasks, with 75 training examples being sufficient to achieve a Spearman's $\rho > 0.30$ on dialogue and open-ended generation while retaining above 60\% coverage. As more data is used, both coverage and the Spearman correlation improve, though there are diminishing returns. Unsurprisingly, the variation in performance across random subsets also drops as more data is used.

\begin{table}
    \centering
    \small
    \begin{tabular}{lccc}
\toprule
Model & Train Topic & $\rho$ & Coverage \\
\midrule

ROUGE-L & 3 & \ \ \ 0.441\ & 1.0 \\
WMS (ELMo) & 3 & \ \ \ 0.443\ & 1.0 \\
SMS (ELMo) & 3 & \ \ \ 0.451\ & 1.0 \\
S+WMS (ELMo) & 3 & \ \ \ 0.490\ & 1.0 \\ \midrule

& 1   &   \ \ \ 0.301\  &  1.0 \\
& 2   &   \ \ \ 0.392\  &  1.0 \\
& 4   &  $-$0.128 &  1.0 \\
& 5   &   \ \ \ \ \ 0.466* &  1.0 \\
BLEU Neighbors & 6 & \ \ \ 0.383\  & 1.0 \\
& 7   &   \ \ \ 0.305\ &  1.0 \\
& 8   &   \ \ \ \ \  \bf 0.500* &  1.0 \\
& 9   &   \ \ \ 0.375\  &  1.0 \\
& 10  &   \ \ \ \ \  0.437* &  1.0 \\
& $\widetilde{3}$ &  \ \ \ \ \  0.465* & 1.0 \\

\bottomrule
\end{tabular}
    \caption{Spearman's $\rho$ between predicted essay quality and the ground truth, where the test essays are from Topic \#3 and * denotes $p < 0.01$. When using essays from Topic \#8 as the training data, BLEU Neighbors is state-of-the-art, even beating out models with access to a gold-standard reference essay for Topic \#3.}
    \label{tab:essay_grading}
\end{table}

\subsection{Automated Essay Grading}
\label{ssec:essays}

In Table \ref{tab:essay_grading}, we report the performance on the essay grading task described in Section \ref{sec:experiments}, where the goal is to score essays from Topic \#3 of the ASAP-SAS dataset. Unlike the NLG tasks, every test example here is a multi-sentence paragraph, which makes scoring more difficult: ten random sentences may be high-quality on their own while making little sense when put together. The difficulty of this task is compounded by the fact that the ground-truth quality of each essay is based on a gold-standard reference for Topic \#3. Since BLEU Neighbors does not use references, it is at a disadvantage compared to approaches that do, such as ROUGE-L. Excluding ROUGE-L, all the models we list in Table \ref{tab:essay_grading} are optimal transport methods that leverage text embeddings \citep{clark2019sentence}. 

Despite not being given the gold-standard reference, when BLEU Neighbors is trained with sample essays from Topic \#8, it achieves a new state-of-the-art: a Spearman's $\rho$ of 0.500 between its predicted scores and the ground-truth quality judgments. However, due to the small amount of test data, this improvement over the state-of-the-art is not statistically significant at $p <0.01$ when using a Williams test. Still, its coverage is 100\%, meaning that it makes predictions for all of the test essays. As seen in the second half of Table \ref{tab:essay_grading}, the performance of the model depends strongly on which topic the training data is sourced from. This is unsurprising, given that some topics are more related to \#3 than others. Some topics (e.g., \#4) are so different from the test topic that its training examples are of no use, leading to very poor quality estimates. When we use essays from all topics but topic \#3 as the training data -- denoted in Table \ref{tab:essay_grading} as $\widetilde{3}$ -- we still outperform most of the past approaches.

\section{Limitations and Future Work}
\label{sec:future}

\paragraph{Quality + Diversity} Although the BLEU Neighbors model does not measure language diversity, this is by design. Consider that if an NLG model were ideal, even the optimal discriminator could not tell whether its outputs were human- or machine-generated. \citet{hashimoto2019unifying} proved that such an optimal discriminator would only need two statistics, a measure of language diversity (e.g., perplexity) and a measure of language quality. The former is trivial to compute -- it is the latter that is cost- and time-intensive, and which we thus try to automate using BLEU Neighbors. These two measures can be combined using a metric such as HUSE \citep{hashimoto2019unifying}, meaning that it is sufficient for our model to predict quality alone. The next step would be to use such a hybrid metric in rapidly evaluating NLG models during development.

\paragraph{Preventing ``Hacks''} How can we prevent the NLG model being evaluated from ``hacking'' a BLEU Neighbors model so as to receive inflated quality estimates for all its outputs? As mentioned in section \ref{sec:model}, one way to prevent this is to use disjoint training sets for the NLG model and BLEU Neighbors, so that the former has no idea what the latter considers a high-quality candidate. Additionally, it would help to have a large set of training examples for BLEU Neighbors and then subsample it during each evaluation instance, as that would discourage NLG models from generating $n$-grams that just so happen to occur in one or two high-quality examples in the training data. Moreover, BLEU Neighbors is intended to speed up NLG model development -- not supplant humans -- so any attempts to inflate quality estimates during development would have poor long-term outcomes.

\paragraph{Metric Learning} The success of BLEU Neighbors can largely be ascribed to it using $\text{BLEU}^*$, a variant of the BLEU-4 score, as the kernel function in sentence space. Despite its simplicity, $\text{BLEU}^*$ works surprisingly well. There is likely a more convoluted variant of BLEU-4 that works even better for this purpose -- one that excludes stopwords, one that places greater weight on rarer $n$-grams, etc. Instead of specifying a kernel function, it may also be possible to learn one. For example, instead of representing each sentence as a sequence of words, one could transform it into a sentence embedding and then learn a kernel function as a metric in the embedding space. This is one direction of future work.

\paragraph{Diverse Datasets} Although BLEU Neighbors performs well in our experiments, because of the small size of the datasets we use, not all results are statistically significant. One limitation of the HUSE datasets in particular is that, as mentioned earlier, the annotators scored different subsets of the data. In order to more faithfully compare our method against human annotators, we need larger datasets from a more diverse array of tasks, where every example is scored by every annotator. 

Because of the leave-one-out paradigm we use on the HUSE datasets, the test examples were scored in part with the help of scored training examples that were generated by the same model. Table \ref{tab:cross_task} shows that cross-task performance is generally poor, with the exception of dialogue data. Would the performance still be poor if we used model-generated training examples from the same task but used a different model to generate them? This is a possibility that should be explored. It is also unclear what exactly is driving the success of BLEU Neighbors. For example, if it is exploiting annotation artefacts, then its success would be far less impressive \citep{gururangan2018annotation}. Understanding these possible failure cases is an important direction for future work. Developing a theoretical understanding of BLEU Neighbors -- as has been done with static word embeddings, for example \citep{levy2014neural,ethayarajh2018towards,ethayarajh2019understanding,ethayarajh2019rotate} -- would be ideal.

\section{Conclusion}

The absence of a reference-less evaluation metric for language quality has been an impediment to developing NLG models. To address this problem, we proposed \emph{BLEU Neighbors}, a nearest neighbors model that leverages the BLEU score as a kernel function in sentence space. Our simple approach worked surprisingly well: it outperformed human annotators -- on average -- in predicting the quality of dialogue and open-ended generation data. We also found BLEU Neighbors to be state-of-the-art on automatically grading essays, even beating out models that had access to a gold-standard reference essay. Moreover, our model is fast, data-efficient, and easy-to-use; it has only two hyperparameters and those have settings that work universally well, across various tasks. Still, BLEU Neighbors is intended to complement, not supplant, human evaluation -- its speed, simplicity, and ease of use makes it ideal for rapidly iterating on NLG models long before any human evaluation is done.

\section*{Acknowledgments}

Many thanks to Alex Tamkin and Peng Qi for detailed feedback. We thank Nelson Liu and Tatsunori Hashimoto for helpful discussion. KE is supported by an NSERC PGS-D.

\bibliography{emnlp2020}
\bibliographystyle{acl_natbib}

\end{document}